\documentclass{article}
\usepackage{spconf,graphicx}
\usepackage{amsmath}
\usepackage{mathrsfs}
\usepackage{amsfonts}
\usepackage{amssymb}
\usepackage{algorithm}
\usepackage{algorithmic}
\usepackage{booktabs}\usepackage{multirow}

\title{HRFA: HIGH-RESOLUTION FEATURE-BASED ATTACK}
\name{Zhixing Ye$^{\star}$, Sizhe Chen$^{\star}$, Peidong Zhang, Chengjin Sun, Xiaolin Huang$^{\dag}$
\thanks{$^{\star}$These authors contribute equally.}\thanks{$^{\dag}$Corresponding author.}}
\address{Institute of Image Processing and Pattern Recognition, Shanghai Jiao Tong University, China\\
\{yzx1213, sizhe.chen, cq-zhang-2016, sunchengjin, xiaolinhuang\}@sjtu.edu.cn}

\begin{document}
%
\maketitle
\begin{abstract}

Adversarial attacks have long been developed for revealing the vulnerability of Deep Neural Networks (DNNs) by adding imperceptible perturbations to the input. Most methods generate perturbations like normal noise, which is not interpretable and without semantic meaning. In this paper, we propose High-Resolution Feature-based Attack (HRFA), yielding authentic adversarial examples with up to $1024 \times 1024$ resolution. HRFA exerts attack by modifying the latent feature representation of the image, i.e., the gradients back propagate not only through the victim DNN, but also through the generative model that maps the feature space to the image space. In this way, HRFA generates adversarial examples that are in high-resolution, realistic, noise-free, and hence is able to evade several denoising-based defenses. In the experiment, the effectiveness of HRFA is validated by attacking the object classification and face verification tasks with BigGAN and StyleGAN, respectively. The advantages of HRFA are verified from the high quality, high authenticity, and high attack success rate faced with defenses.
\end{abstract}
\section{Introduction}
\label{sec:intro}
In recent years, deep learning has achieved an extraordinary success in computer vision, natural language processing and other fields \cite{lecun2015deep}. However, because of its black-box characteristics, the robustness and reliability of DNNs have long been suspected. One evidence that DNNs are far from perfect, is the existence of adversarial examples \cite{szegedy2013intriguing, goodfellow2014explaining, tang2019adversarial, chen2020universal, anonymous2021relevance}, where DNNs are easily fooled by elaborately-crafted but imperceptible perturbations on the input and make false predictions at a high confidence.

Generally, adversarial attacks fix the network parameters and optimize the input to maximize a training loss. The current mainstream of adversarial attacks adds perturbations directly in image space \cite{goodfellow2014explaining,chen2020universal, carlini2017towards}, i.e., the attack gradients back propagate to the input image and modify it to become adversarial. In this way, the calculated perturbations are without semantic meanings like random noise, leading to \emph{noise-based attacks}. Though effective, these attacks do not produce authentic adversarial samples, i.e., images in the distribution of natural photos  \cite{prakash2018deflecting}. Thus, noise-based adversarial examples could be easily detected  \cite{liu2019feature}.

\begin{figure}[htbp]
    \centering
    \includegraphics[width=\hsize]{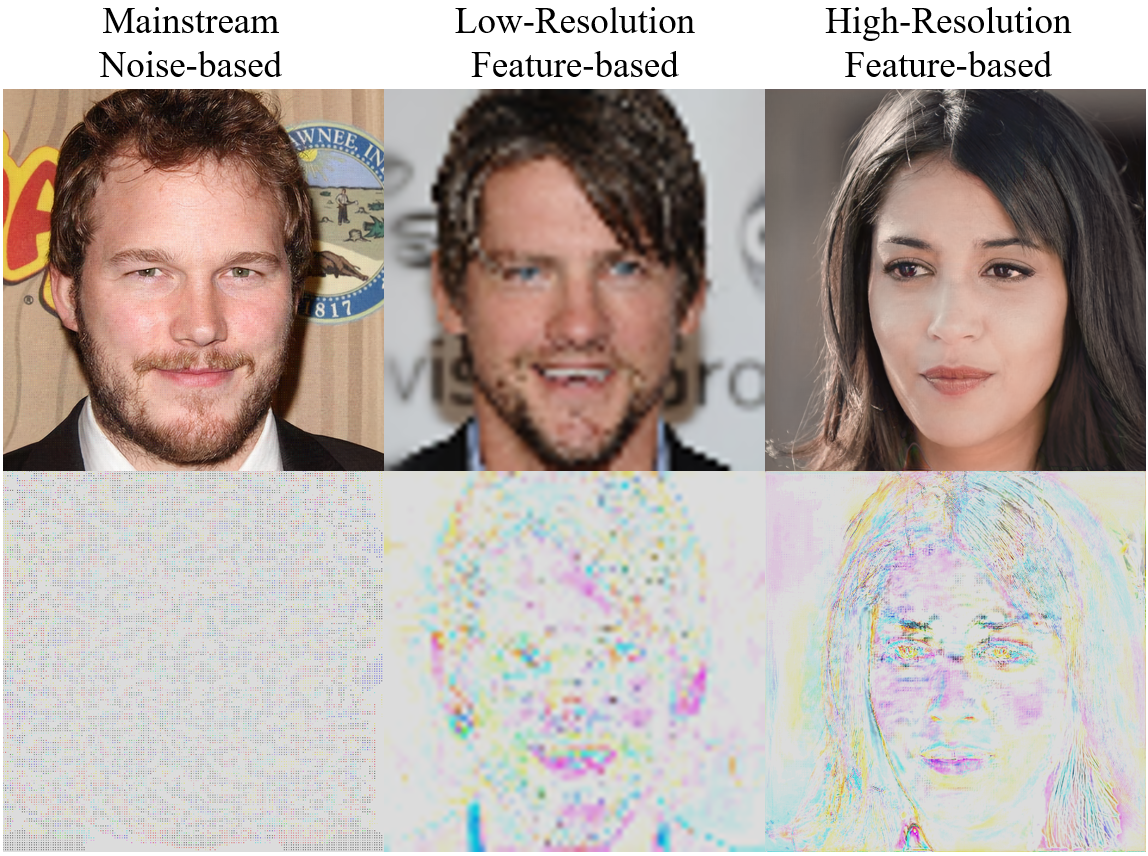}
    \caption{Comparison among the mainstream noise-based attacks (e.g., PGD \cite{madry2017towards} here), low-resolution feature-based attack  (e.g., \cite{song2018constructing} here) and our high-resolution feature-based attack from left to right. The top row is adversarial examples, and the bottom row is the corresponding perturbations, which are equally multiplied for a clear presentation.}
    \label{fig:intro}
\end{figure}

Inspired by our Type I attack \cite{tang2019adversarial} and AoA attack \cite{chen2020universal}, it will be more interesting to link the attack perturbations with semantic meanings. In contrast to noise-based attacks, \emph{feature-based attacks} produce interpretable perturbations, which could not be inhibited by defense methods like noise-based attacks. As shown in Fig. \ref{fig:intro}, perturbations of our high-resolution feature-based method capture the outline of the object and highlight the most sensitive regions of the image, e.g., the leap and eyes, revealing weaknesses of the networks in a structured and explainable manner. 
However, previous feature-based attacks \cite{wang2019gan, song2018constructing} can only generate low-resolution examples as in the middle column of Fig.\ref{fig:intro}, which is significantly not authentic compared to our attack in the right column.

In this paper, we propose High-Resolution Feature-based Attack (HRFA). HRFA adopts the image loss to guarantee the imperceptibility of input and the net loss to mislead the victim DNN. We optimize the total loss by modifying the latent feature representation of the image, i.e., the gradients back propagate not only through the victim DNN, but also through the generative model that maps the feature space to the image space. Because we directly manipulate the low-dimensional latent feature map to attack, the resulting image perturbations are continuous and interpretable. HRFA is a universal attack framework applicable to various tasks. We conduct comprehensive experiments to attack classification and face verification with BigGAN \cite{brock2018large} and StyleGAN \cite{Karras_2019_CVPR}, respectively. Results show that HRFA produces authentic adversarial examples with resolution up to $1024 \times 1024$. Furthermore, they are immune to several defenses compared to existing attacks.
\section{Method}
\label{sec:print}

In this section, we formulate the adversarial attack problem, and then describes the general idea of HRFA, followed by its detailed procedures.

Consider a deep neural network $f:\mathcal{X}\rightarrow \mathcal{Y}$ taking an image $x \in \mathcal{X}$ as input, and outputs a prediction $f(x)\in \mathcal{Y}$, where $f(x)$ can be a continuous value or a discrete label, according to specific tasks. The goal of adversarial attack is to generate an adversarial example $x'$ so that $f(x')$ gives a incorrect prediction. Meanwhile, $x'$ is very close to $x$ in the human's eyes.

In order to perform the mappings from the image space to the feature space, a parameterized generative model is needed. Generative Adversarial Network (GAN) and Variational Auto-Encoder (VAE) are common generative models, and GAN has an extreme fidelity and hence is chosen in HRFA. A GAN is composed of a generator $g(z)$ and a discriminator $d(x)$, where $z \in \mathcal{Z} \subset \mathbb{R}^{n}$ denotes an $n$-dimension vector in the feature space $\mathcal{Z}$, $x \in \mathcal{X} \subset \mathbb{R}^{m}$  denotes an $m$-dimension squeezed vector in the image space $\mathcal{X}$. A trained generator maps a source of noise $z$ to a synthetic image $x=g(z)$. Since $n\ll m$, the latent vector $z$ contains information in a much condensed manner, which is named \textit{feature} in our paper as in \cite{xiao2018generating}.

\begin{figure}[htbp]
  \centering
  \includegraphics[width=\hsize]{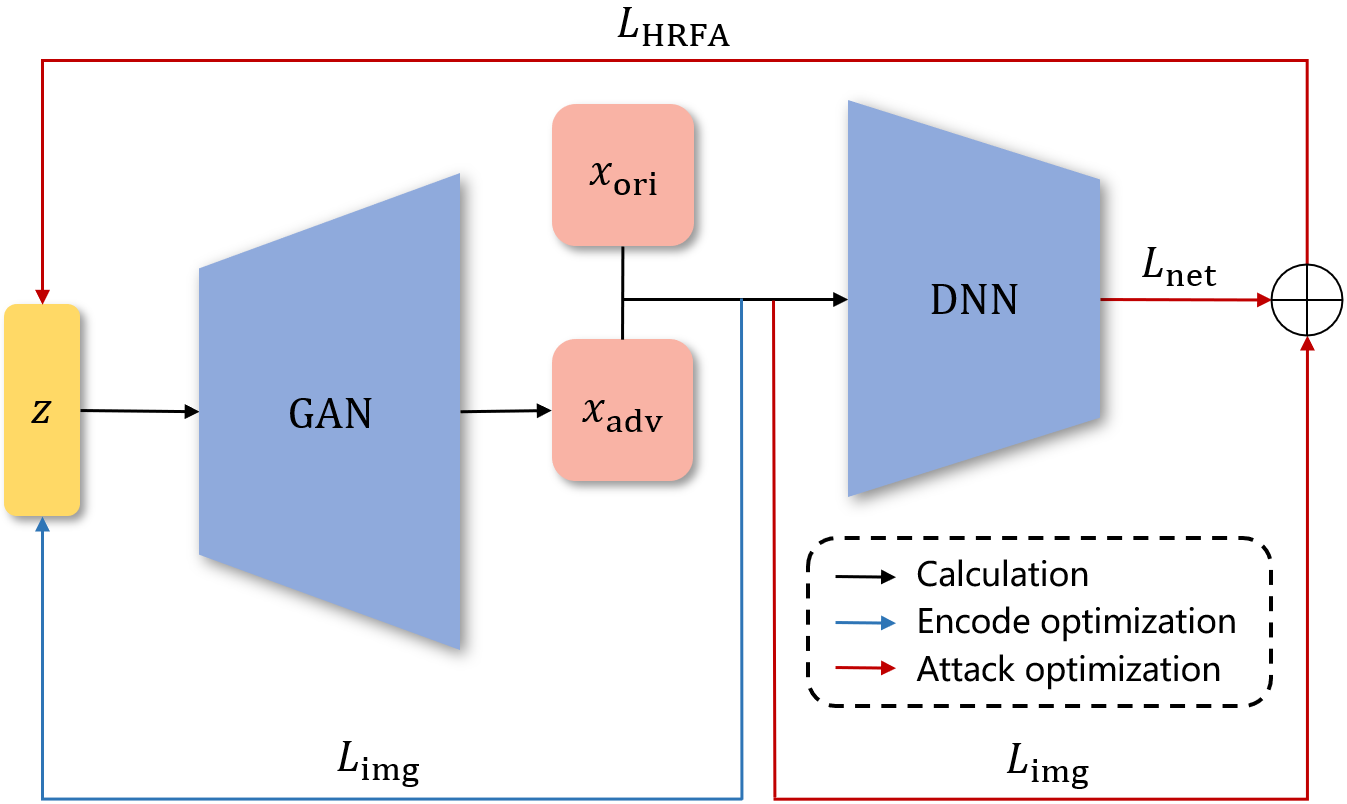}
\caption{Framework of HRFA. HRFA adopts a generator $g$ to craft adversarial examples to the victim neural network $f$. Both the generator $g$ and net $f$ are pretrained network with fixed parameters. $g$ takes a feature vector $z$ as the input and generates an image $x=g(z)$, which is sent to the victim network $f$ for a prediction $y=f(x)$. HRFA encodes and attacks by modifying latent feature $z$.}
\label{fig:framework}
\end{figure}

Fig. \ref{fig:framework} illustrates the overall architecture of HRFA. Firstly, a latent feature $z$ is randomly initialized to generate an original example $x_\mathrm{ori}=g(z)$. Then, we compute the loss and back propagate it not only through the victim network, but also through the generator to $z$, which is then optimized. The attack loss is $L_{\mathrm{HRFA}}=L_{\mathrm{img}}+\lambda L_{\mathrm{net}}$, where $\lambda$ is a hyperparameter to weight two terms. The first component
\begin{equation}\label{l_img}
    L_{\mathrm{img}}\triangleq \| x_{\mathrm{adv}}-x_{\mathrm{ori}}\|_2
\end{equation}
is named \textit{image loss}, which denotes the $\ell_2$-norm distance between the clean example and the adversarial example. The image loss guarantees the similarity between them during the attack. The second component
\begin{equation}\label{L_net}
    L_{\mathrm{net}}\triangleq L(f, x),
\end{equation}
named \textit{net loss}, misleads the victim network $f$ to make false predictions. Note that HRFA framework is applicable in attacking various tasks, so the choice of net loss depends on the victim model and would be specified in experiments given the tasks. With the calculated loss, we optimize as
\begin{equation}
    z = z - \Gamma(\frac{\partial L_\mathrm{HRFA}}{\partial z}).
\end{equation}
The gradients back propagate through the neural network, and then through the generator to the feature space. The gradients of HRFA loss determine the optimization direction of the feature vector to minimize the loss. The optimization is repeated until $f(g(z))$ gives a incorrect prediction or exceeds a certain threshold. Since we modify the feature in low-dimensional latent space, the resulting perturbations in high-resolution image space is much more interpretable.

In the descriptions above, the adversarial example is crafted from a virtual synthesized clean image, which is generated from the randomly initialized $z$. Actually, attack from a certain original example is feasible if an encoder is available or the generator (decoder) $g$ is reversible. It has been proved in \cite{aberdam2020and} that a well-trained generative model can be reversed using gradient descent. Specifically, we randomly initialize a feature vector $z$. By minimizing the image loss $L_{\mathrm{img}}$ in (\ref{l_img}) and optimize $z$, we can finally reach $z^{\star}$ satisfying $x_\mathrm{ori}=g(z^{\star})$, thus realizing the mutual conversion between feature space and image space. This process is named \emph{encode optimization} before the \emph{attack optimization} described above. Algorithm \ref{HRFA} illustrates the pseudo code of HRFA.
\begin{algorithm}[htbp]
\caption{HRFA}
\label{HRFA}
\textbf{Input}: victim network $f$, prediction loss $L(f, x)$, generator $g$, original sample $x_\mathrm{ori}$, encoding loss threshold $\tau_\mathrm{encode}$, attacking loss threshold $\tau_\mathrm{attack}$, loss trade-off term $\lambda$, maximum attack iteration $m$, and Adam learning rate $lr$. \\
\textbf{Output}: adversarial sample $x$\\
\begin{algorithmic}[1]
\STATE \# Encode optimization
\STATE Let $z=\mathrm{random\_normal()}$, $L_\mathrm{img} = \|g(z)-x_\mathrm{ori}\|_2$
\WHILE{$L_\mathrm{img} \geq \tau_\mathrm{encode}$}
    \STATE $x = g(z)$
    \STATE $L_\mathrm{img} = \|x-x_\mathrm{ori}\|_2$
    \STATE $z = z - \Gamma(\frac{\partial L_\mathrm{img}}{\partial z})$
\ENDWHILE
\STATE
\STATE \# Attack optimization
\STATE Let $iter=0$
\WHILE{$iter < m$ \AND $L_\mathrm{net} \geq \tau_\mathrm{attack}$}
    \STATE $x = g(z)$
    \STATE $L_\mathrm{img} = \|x-x_\mathrm{ori}\|_2$
    \STATE $L_\mathrm{HRFA}=L_\mathrm{img}+\lambda L_\mathrm{net}$
    \STATE $z = z - \Gamma(\frac{\partial L_\mathrm{HRFA}}{\partial z})$
    \STATE $iter = iter + 1$
\ENDWHILE
\STATE \textbf{return} $x$
\end{algorithmic}
\end{algorithm}

Note that HRFA is a universal framework, which is applicable to different tasks only if $L_{\mathrm{net}}$ is specified. For instance, in a classification task, the net loss can be specified as the confidence of predicting $x$ as its original label $y$. In a face verification task, it can be defined as the predicted perceptual distance between two faces.

\section{EXPERIMENTS}
\label{sec:experiment}
In this section, we conduct comprehensive experiments on both object classification and face verification. Since HRFA adversarial examples are high-resolution and authentic, it is intuitive that such adversarial examples could evade defenses, which are for traditional unrealistic perturbations. Therefore, we compare our method with mainstream attacks in defense evasion rate, which denotes the error rate of victim model on the adversarial examples processed by defenses. Experiments on object classification and face verification are implemented on PyTorch \cite{paszke2019pytorch} and TensorFlow \cite{abadi2016tensorflow} with 4 NVIDIA GeForce RTX 2080Ti GPUs.

\subsection{HRFA on object classification}\label{sec:biggan}
The object classification is conducted on ImageNet \cite{deng2009imagenet}, a large-scale benchmark commonly used in attacks. BigGAN \cite{brock2018large} ($512 \times 512$ output) is a high-fidelity generator on ImageNet and thus chosen here. Popular ResNet152 \cite{he2016deep} and DenseNet121 \cite{huang2017densely} are victim models. In classification, $L_{\mathrm{net}}$ is the confidence of $f$ on the correct label class. 
Two mainstream attacks, PGD \cite{madry2017towards} and C\&W \cite{carlini2017towards}, are chosen as baselines. We randomly craft 2K adversarial examples for class ``goldfinch'' and ``bald eagle'', which the BigGAN is more capable of generating. The hyperparameters in Algorithm \ref{HRFA} are set as $\lambda=50, m=100, lr=0.05$, and the step size of baseline is set as $\alpha=0.003$. All attacks stop when the victim model classifies incorrectly.
\begin{figure}[htbp]
    \centering
    \includegraphics[width=\hsize]{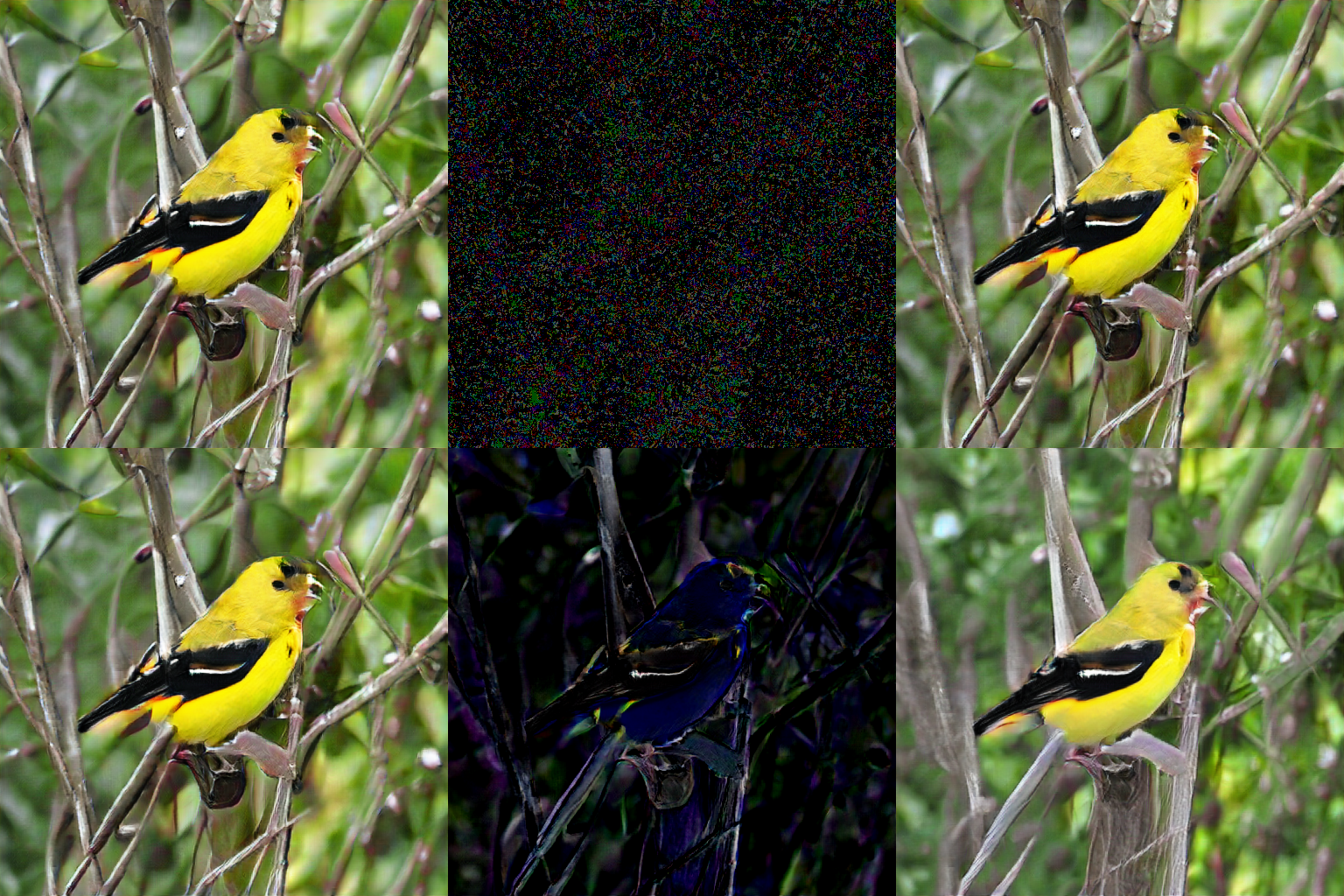}
    \caption{A randomly chosen example for object classification attacked by PGD (top row) and HRFA (bottom row). Column 1-3 stands for the original example, perturbations, and adversarial example, from left to right.}
    \label{fig:biggan}
\end{figure}

Fig. \ref{fig:biggan} demonstrates an adversarial example generated by PGD and HRFA, from which we can observe that HRFA perturbation is like a semantic change in the object or background rather than a noise-like perturbation in image space. That is to say, the perturbation is seemingly large, but has subtle perceptual influence in the human's eye, and thus with high authenticity, which leads to HRFA's immune to denoising-based defenses, as shown in Table \ref{tab:biggan}. In terms of defense methods, we choose Mean Filter with $\sigma=5$, Pixel Deflection \cite{prakash2018deflecting} and JPEG compression \cite{liu2019feature} with hyperparameters suggested by their inventors given their effectiveness in preventing mainstream attacks. All these defenses take a potential adversarial example as input and outputs a processed image, which is aimed to be predicted correctly by the victim model. The evasion rate is defined as $N_{\mathrm{err}}/N$, where $N$ is the total number of processed images, and $N_{\mathrm{err}}$ denotes the number of incorrectly-predicted processed images. The higher the evasion rate, the more aggressive the adversary is.

\begin{table}[tbp]
\caption{Defense evasion rates for object classification. MF, PD, and JPEG stand for using Mean Filter, Pixel Deflection, and JPEG compression defenses, respectively.\\}
\resizebox{\linewidth}{!}{
\begin{tabular}{ccccccccc}
\toprule
\multirow{2}{*}{Method} & \multicolumn{3}{c}{DenseNet201} & \multicolumn{3}{c}{ResNet152} \\
\cmidrule(r){2-4} \cmidrule(r){5-7}
&  MF   &  PD  &  JPEG   & MF   &  PD  &  JPEG
\\
\midrule
HRFA & 99.56 & 97.47 & 91.67 & 100.00 & 99.85 & 99.95\\
PGD  & 43.61 & 52.88 & 91.70 & 76.60 & 43.63 & 57.29\\
C\&W & 38.45 & 31.54 & 40.31 & 71.89 & 24.67 & 23.92\\
\bottomrule
\label{tab:biggan}
\end{tabular}}
\end{table}

As shown in Table \ref{tab:biggan}, HRFA outperforms baselines in defense evasion rate to a large extent, with success rates higher than 90\%, regardless of the victim model and defense methods. This suggests that HRFA adversarial examples are of high reality and are immune to denoising-based defenses.

\subsection{HRFA on face verification}
In order to show the generalization of our method, we also conduct face verification task on CelebA-HQ \cite{abadi2016tensorflow} with StyleGAN \cite{karras2017progressive} ($1024 \times 1024$ output) as the generative model and FaceNet \cite{schroff2015facenet} as the victim model. Since StyleGAN is well-trained and thus reversible, we implement our framework based on real images from CelebA-HQ, which is a dataset containing 30K $1024 \times 1024$ images of celebrity faces. 3K images are randomly chosen to generate adversarial examples and evaluate the performance.

Different from classification task, face verification model extracts features from two faces, and outputs a value $d_{face}$ denoting the face distance. These two faces are predicted as one person if $d_{face}\leq\tau$, where $\tau$ is a threshold decided according to the actual application. We set $\tau=1.1$ in this experiment according to \cite{schroff2015facenet}, and the goal of the adversary is to increase the face distance between original images and adversarial ones (crafted from the original ones) to $\tau$. Consequently, the net loss can be specified as $L_{\mathrm{net}}=f(x, x_{\mathrm{adv}})$, because a greater value means a higher confidence of a correct answer, and we aim to lower it during optimization. We compare the performance of HRFA with PGD (without C\&W because it is only applicable in classification). The hyperparameters in Algorithm \ref{HRFA} are set as $\lambda=10, \tau_\mathrm{encode}=10, \tau_\mathrm{attack}=1.6, m=1000$, and the step size of baseline is set as $\alpha=2$.

Here, we also compare the defense evasion rate for different attacks. Besides that, We also report the average face distance $d_{face}$ in brackets in Table \ref{tab:stylegan} to clearly report the aggression of attacks. Because in face verification, judging the adversarial example as a different person, i.e., $d_{face}\geq\tau$, indicates a mistake. Given that HRFA on face verification includes encoding procedure, we also want to measure the distance between clean examples and adversarial examples. In our experiments, we evaluate the authenticity of adversarial examples numerically by Structural SIMilarity (SSIM) \cite{wang2004image}, because it is a perception-based metric, revealing the structure distance of the objects in the visual scene, which is consistent with our requirements in assessing the perceptual difference.

\begin{table}[!ht]
\caption{Defense evasion rates, face distances and SSIMs for face verification. A high SSIM indicates a perceptual similarity between original examples and adversarial ones.\\}
\centering
\label{Tab01}
\resizebox{\linewidth}{!}{
\begin{tabular}{ccccc}
\toprule
\multirow{2}{*}{Method} & \multirow{2}{*}{SSIM} & \multicolumn{3}{c}{Defense Evasion Rate (Face Distance)} \\
\cmidrule(r){3-5}
&& MF & PD   &  JPEG
\\
\midrule
HRFA  &0.757    &93.97 (1.35)   & 58.20 (1.14) & 99.57 (1.52)\\
PGD   &0.465  & 6.34 (0.78)   & 12.60 (0.77) &94.93 (1.44) \\
\bottomrule
\end{tabular}
}
\label{tab:stylegan}
\end{table}

\begin{figure}[htbp]
    \centering
    \includegraphics[width=.95\hsize]{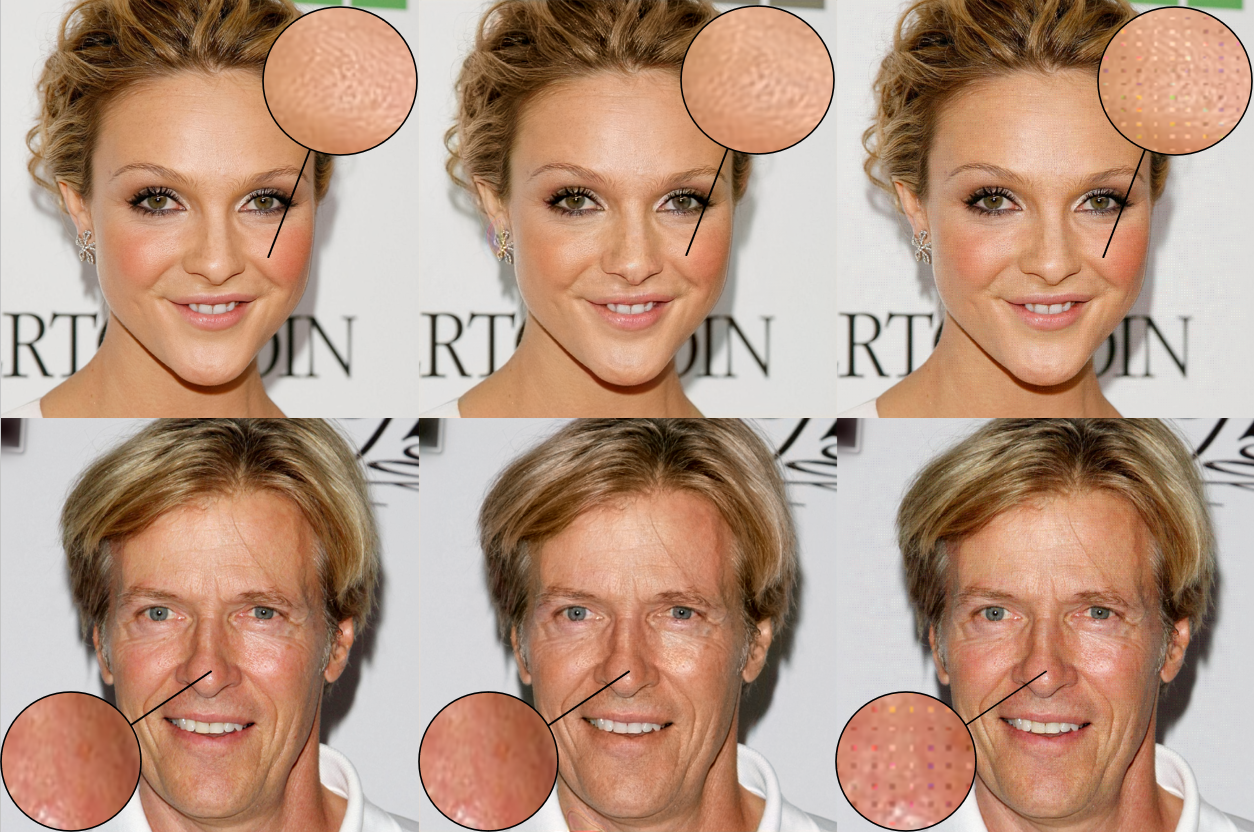}
    \caption{Two examples of adversarial attack on face verification. Column 1-3 stands for the clean example, HRFA adversarial examples, PGD adversarial examples, respectively. Some areas are resized to observe the details.}
    \label{fig:stylegan}
\end{figure}

As shown in Table \ref{tab:stylegan}, the SSIM of adversarial examples generated by HRFA are 60\% higher than PGD on average. This is because HRFA generates perturbations concentrating on the edge of the object, which are more natural with semantic meaning. By contrast, the perturbations generated by PGD are like random noise. In terms of defense, HRFA also obviously outperforms the mainstream PGD by up to 15 times. From a closer observation in Fig. \ref{fig:stylegan}, HRFA perturbations are too subtle to be perceived as PGD's, which means they are less prone to be identified as high-frequency noise and eliminated by defense methods.

\section{CONCLUSION}
\label{sec:conclusion}
In this paper, we propose an attack framework named High-Resolution Feature-based Attack (HRFA). We generate high-resolution (up to $1024 \times 1024$ resolution) adversarial examples by varying their latent feature, i.e., back propagating the gradients also through the generative model. By comprehensive experiments on object classification and face verification, we compare HRFA with mainstream attacks faced with several defenses. Results show that HRFA outperforms baseline attacks to a large extent in terms of adversarial examples' quality and defense evasion rate, indicating the effectiveness of HRFA on various attack scenarios and tasks.

\clearpage
\fontsize{9.0pt}{\baselineskip}\selectfont
\bibliographystyle{IEEEbib}
\bibliography{refs}

\end{document}